\documentclass[letterpaper, 10 pt, conference]{ieeeconf}  

\IEEEoverridecommandlockouts                              

\usepackage{graphics} 
\usepackage{epsfig} 
\usepackage{mathptmx} 
\usepackage{times} 
\usepackage{amsmath} 
\usepackage{amssymb}  
\usepackage{float}
\usepackage{subfigure}
\usepackage{caption, copyrightbox}
\usepackage{color}
\usepackage{bbm}
\usepackage{url}

\makeatletter
\renewcommand{\@thesubfigure}{\hskip\subfiglabelskip}
\makeatother

\definecolor{white}{RGB}{255,255,255}
\definecolor{red}{RGB}{255,0,0}

\title{PENet: Towards Precise and Efficient Image Guided Depth Completion}

\author{Mu Hu$^{1}$, Shuling Wang$^{1}$, Bin Li$^{1}$, Shiyu Ning$^{2}$, Li Fan$^{2}$ and Xiaojin Gong$^{1*}$
\thanks{$^{1}$M. Hu, S. Wang, B. Li, and X. Gong are with the College of Information Science and Electronic Engineering, Zhejiang University, Hangzhou, China. X. Gong is the corresponding author. {\tt\small [muhu, 11831041, 3130102392, gongxj]@zju.edu.cn}}
\thanks{
$^{2}$S. Ning and L. Fan are with the Department of Turing Solution, Hisilicon, Huawei Shanghai, China. {\tt\small ningshiyu@hisilicon.com, fanli11@huawei.com}}%
}

\begin{document}

\maketitle
\thispagestyle{empty}
\pagestyle{empty}

\begin{abstract}
Image guided depth completion is the task of generating a dense depth map from a sparse depth map and a high quality image. In this task, how to  fuse the color and depth modalities plays an important role in achieving good performance. This paper proposes a two-branch backbone that consists of a color-dominant branch and a depth-dominant branch to exploit and fuse two modalities thoroughly. More specifically, one branch inputs a color image and a sparse depth map to predict a dense depth map. The other branch takes as inputs the sparse depth map and the previously predicted depth map, and outputs a dense depth map as well. The depth maps predicted from two branches are complimentary to each other and therefore they are adaptively fused. In addition, we also propose a simple geometric convolutional layer to encode 3D geometric cues. The geometric encoded backbone conducts the fusion of different modalities at multiple stages, leading to good depth completion results. We further implement a dilated and accelerated CSPN++ to refine the fused depth map efficiently. The proposed full model ranks 1st in the KITTI depth completion online leaderboard at the time of submission. It also infers much faster than most of the top ranked methods. The code of this work is available at \textcolor{red}{\url{https://github.com/JUGGHM/PENet_ICRA2021}}. 

%
%

%
%

\end{abstract}

\section{Introduction}
Image guided depth completion aims to predict a dense depth map from a sparse one with the guidance of a high-resolution color image. This task has been attracting considerable research interest due to its importance in various computer vision applications, such as autonomous driving, 3D reconstruction, and augmented reality. A sparse depth map is usually obtained by projecting 3D point clouds collected by ranging sensors like LiDARs in outdoor environments. However, even if a high-end LiDAR is employed, the projected depth maps are still highly sparse and also noisy around object boundaries. These defects make depth completion a challenging problem. 

To address this problem, a wide variety of methods have been developed. Recent approaches are mainly based on deep convolutional neural networks. Considering that color and depth are two different modalities, most previous methods adopt two-branch network architectures in order to fuse the two modalities. For instance, Jaritz et al.~\cite{Jaritz2018semantic} and Hua et al.~\cite{Hua2018normalized} use two encoders to extract features from each modality separately and then fuse them into one decoder. Tang et al.~\cite{Tang2019guided} construct two encoder-decoder networks to extract color and depth features and take a decoder-encoder fusion scheme. In these networks, each branch inputs only one modality and therefore only late fusion is considered. 

Two-branch architectures are also constructed in some works, such as FusionNet~\cite{Gansbeke2019} and DeepLiDAR~\cite{Qiu2019deeplidar}, to perform both early and late fusion. FusionNet~\cite{Gansbeke2019} consists of two branches to extract local and global information respectively. DeepLiDAR~\cite{Qiu2019deeplidar} is a network composed of a color pathway and a surface normal pathway. In these networks, each branch takes two modalities as inputs and the multi-modality fusion is performed at multiple stages. By this means, better fusion can be achieved, which further results in better depth completion performance. However, these two methods~\cite{Gansbeke2019,Qiu2019deeplidar} require extra datasets, such as Cityscapes~\cite{Cordts2016} or synthetic data~\cite{Qiu2019deeplidar}, to pretrain their networks.


%
%
%

Inspired by above-mentioned methods, our work constructs a two-branch network, which consists of a color-dominant (CD) branch and a depth-dominant (DD) branch, as the backbone. In contrast to FusionNet~\cite{Gansbeke2019} and DeepLiDAR~\cite{Qiu2019deeplidar}, we design the branches for different purposes. More specifically, the CD branch aims to extract color-dominant information for depth prediction. It inputs a color image and a sparse depth map and produces a dense depth map. Since this branch is color-dominant, the predicted depth map is relatively reliable around object boundaries but may be too sensitive to the change of color or texture. The DD branch takes as inputs a sparse depth map and the CD depth prediction to produce a dense depth map, which overall is reliable but suffered from the heavy noise existing near object boundaries in the sparse input. It means that the depth maps predicted from two branches are complementary to each other. Therefore, we adaptively fuse them with learned confidence weights. This backbone is able to exploit and fuse color and depth modalities thoroughly. It can also be trained from scratch without using extra datasets.  


In addition, we also propose a simple geometric convolutional layer to encode 3D geometric cues. It simply augments a convolutional layer via concatenating a 3D position map to the layer's input. Assisted with this geometric encoding scheme, our backbone achieves promising performance. Considering that the accurate depth values from the sparse input may not be preserved after prediction, we additionally integrate a module based on CSPN++~\cite{Cheng2020CSPN++} to refine the depth map predicted by our backbone. We design a dilated and accelerated implementation of CSPN++ to make the refinement more effective and efficient.

\begin{figure*}[t]
\centering
\includegraphics[width=0.95\textwidth]{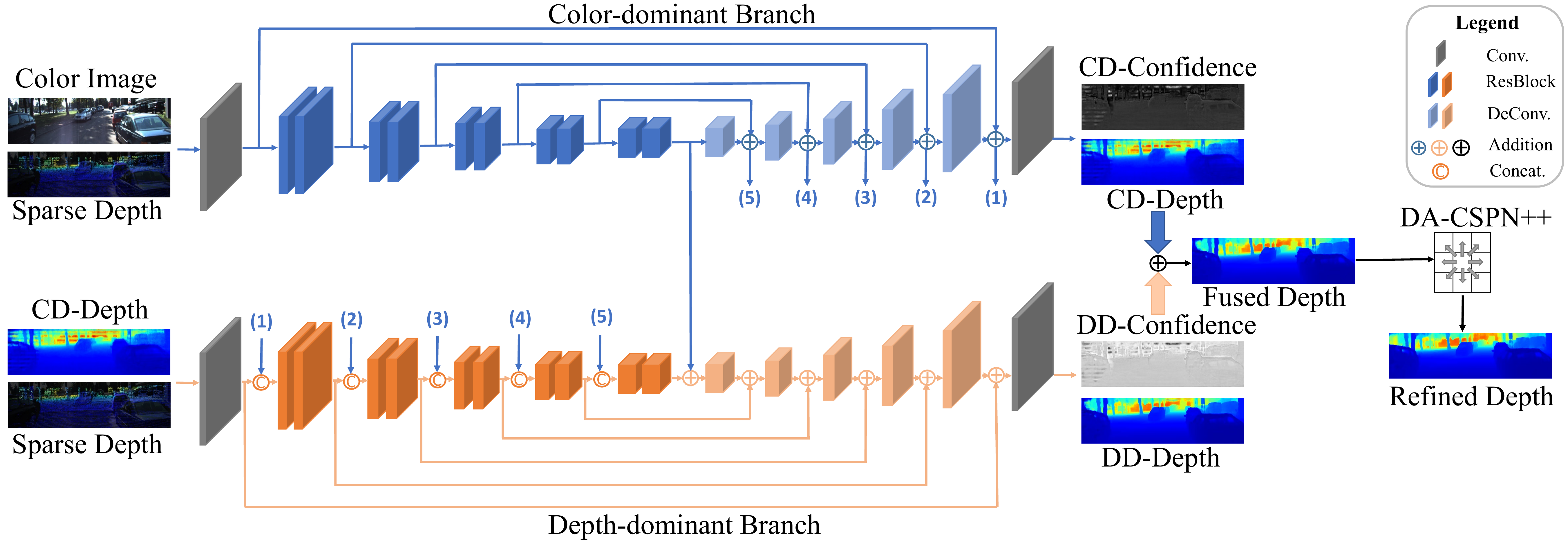} 
\caption{An overview of the proposed framework. It consists of a two-branch backbone and a depth refinement module. The branches predict two dense depth maps, denoted as CD-Depth and DD-Depth respectively, from color-dominant and depth-dominant information. CD-Depth and DD-Depth are adaptively fused and further refined by a dilated and accelerated (DA) CSPN++. Here (1)-(5) denotes multi-scale CD-features which are then concatenated with DD-features.}
\label{fig_framework}
\end{figure*}

The main contributions of our work are summarized as follows:
\begin{itemize}
	\item We construct a two-branch backbone that produces dense depth prediction via exploiting color- and depth-dominant information, respectively, from two branches. This backbone is able to exploit and fuse color and depth modalities thoroughly. 
	\item We propose a geometric convolutional layer to simply encode 3D geometric cues. The geometric encoded backbone outperforms most top ranked and peer-reviewed methods.
	
	\item We design an implementation way to accelerate the depth refinement technique CSPN++, making it much more efficient.
	\item The proposed full model ranks 1st in the KITTI depth completion online leaderboard\footnote{http://www.cvlibs.net/datasets/kitti/eval\_depth.php?benchmark\\=depth\_completion} at the time of submission. Moreover, it infers much more efficiently than most of the top ranked methods. 
	\end{itemize}




\section{Related Work}
\subsection{Depth Completion}
Depth completion aims to produce a dense depth map by completing a sparse depth map, without~\cite{Uhrig2017sparsity,Eldesokey2020uncertainty} or with the guidance of a reference image~\cite{Liu2013,Hua2018normalized,Cheng2018CSPN,Qiu2019deeplidar}. The latter takes advantage of structure information from the guidance image to boost performance and therefore attracts more research interests. The image guided depth completion task has specific challenges, including 1) the input depth map is irregularly sparse and noisy; 2) the color image and the depth map are two different modalities. To address these issues, different sparse invariant convolutions~\cite{Uhrig2017sparsity,Hua2018normalized,Eldesokey2018,Huang2019hms-net}, uncertainty exploration~\cite{Eldesokey2020uncertainty,Gansbeke2019} and multi-modality fusion strategies~\cite{Tang2019guided} have been developed. Besides, various recent methods also exploit multi-scale features~\cite{Eigen2014,Huang2019hms-net,Wang2018multiscale,Li2020MSG}, surface normal~\cite{Qiu2019deeplidar,Xu2019}, semantic information~\cite{Jaritz2018semantic,Schneider2016semantic}, or context affinity~\cite{Cheng2018CSPN,Cheng2020CSPN++,Park2020NLSPN} to improve performance further. Among these methods, we take a two-branch architecture similar to~\cite{Qiu2019deeplidar,Gansbeke2019} as our backbone. But we construct our branches for different purposes and our network is more effective.

\subsection{Geometric Encoding}
As pointed out in~\cite{Chen20192d3d}, 3D geometric clues are important for depth completion. So far various strategies have been developed to encode geometric cues. For instance, UberATG~\cite{Chen20192d3d} applies continuous convolution on 3D points, ACMNet~\cite{Zhao2020context} exploits the graph propagation, DeepLiDAR~\cite{Qiu2019deeplidar} and PwP~\cite{Xu2019} use surface normal to introduce geometric constraints. These methods are either complicated in computation or need extra data for learning. In this work, we propose a geometric convolutional layer to encode 3D geometric cues simply. Our method is inspired by CoordConv~\cite{Liu2018Coordconv}, which encodes 2D position information by simply augmenting an input of a convolution with extra coordinate channels. CoordConv~\cite{Liu2018Coordconv} has demonstrated its effectiveness in position-sensitive applications such as object segmentation~\cite{Wang2020SOLO} and semantic segmentation~\cite{Choi2020fly}. Our experiments show that the proposed geometric convolutional layer can considerably improve the depth completion performance but CoordConv~\cite{Liu2018Coordconv} is not helpful.

\subsection{Spatial Propagation Networks}
The spatial propagation network (SPN) is proposed by Liu et al.~\cite{Liu2014SPN} to learn local affinities that can be exploited in various high-level vision tasks. However, it propagates in a column-wise and row-wise manner, which is inefficient. Cheng et al.~\cite{Cheng2018CSPN} thereby propose a convolutional spatial propagation network (CSPN) for efficiency and meanwhile apply it to refine depth completion results. These two methods perform propagation within a fixed local neighborhood. To dynamically learn the convolutional kernels, CSPN++~\cite{Cheng2020CSPN++} and NLSPN~\cite{Park2020NLSPN} are proposed very recently. The former adaptively learns the convolutional kernel size and iteration number for propagation, while the latter learns deformable kernels. These SPN methods are effective to refine depth predictions but still not so efficient. We adopt CSPN++~\cite{Cheng2020CSPN++} for our depth refinement, but we introduce a dilation scheme to enlarge the neighborhoods and implement the propagation in a much more efficient way.

\section{Methodology}
We design an end-to-end learning framework for image guided depth completion. As shown in Figure~\ref{fig_framework}, the entire framework consists of a two-branch backbone and a depth refinement module. In the backbone, one branch is color-dominant, which predicts a dense depth map mainly relying on color information. The other is depth-dominant, which also predicts a dense depth map but depending more on depth information. The depth maps predicted from two branchs are adaptively fused with learned confidence weights. The fused map is further fed into the refinement module to enhance the depth quality. In this module, we adopt the CSPN++~\cite{Cheng2020CSPN++} technique but make it more effective and efficient via a dilated and accelerated implementation.

\subsection{The Two-branch Backbone}
The two-branch backbone is designed to thoroughly exploit color-dominant and depth-dominant information from their respective branches and make the fusion of two modalities effective. To this end, we build similar encoder-decoder networks in two branches to perform a color-dominant depth prediction and a depth-dominant depth prediction. 

%


%
%
%
%
%
%
%
%
%
%

%

\textit{The color-dominant branch} initially aims to predict a dense depth map from a color image. For the purpose of effectiveness, an aligned sparse depth map is also input to assist depth prediction. In this branch, we build an encoder-decoder network with symmetric skip connections. The encoder contains one convolution layer and ten basic residual blocks, i.e. ResBlocks~\cite{he2016deep}. The decoder has five deconvolution layers and one convolution layer. Each of all convolutional layers are followed by a BN layer and a ReLU activation. Although both a color image and a sparse depth map are input, this branch extracts color-dominant features for depth prediction so that the depth around object boundaries can be learned by taking advantage of structure information in the color image.

\textit{The depth-dominant branch} initially aims to predict a dense depth map by upsampling a sparse one. In this branch, a similar encoder-decoder network is constructed. We additionally adopt a decoder-encoder fusion strategy~\cite{Tang2019guided} to fuse the color-dominant features into this branch. Specifically, the decoder features of the color-dominant branch are concatenated with the corresponding encoder features in the depth-dominant branch. In addition, the depth prediction result obtained from the CD branch is also input to this branch. By this means, the features of color and depth modalities are fused at multiple stages. 


%
%
%

%
%
%
%
%

\textit{Depth fusion.} As two dense depth maps are predicted, we fuse them by following the same strategy in FusionNet~\cite{Gansbeke2019}. Formally, we denote the depth maps obtained from two branches by $\hat{D}_{cd}$ and $\hat{D}_{dd}$ respectively, and the confidence maps by $C_{cd}$ and $C_{dd}$. The fused depth map is obtained by
\begin{equation}
\hat{D}_{f}(u,v) = \frac{e^{C_{cd}(u,v)}\cdot \hat{D}_{cd}(u,v) + e^{C_{dd}(u,v)}\cdot \hat{D}_{dd}(u,v) }{e^{C_{cd}(u,v)} + e^{C_{dd}(u,v)}},
\end{equation}
in which $(u, v)$ denotes a pixel. 



\begin{figure}[t]
	\centering
	\subfigure[(a) A convolutional layer]{\includegraphics[width=0.2\textwidth]{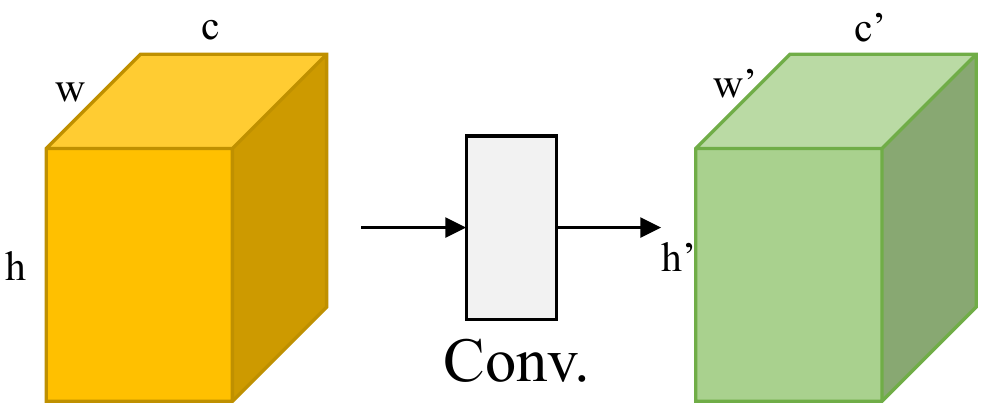}}
	\subfigure[(b) A geometric convolutional layer]{\includegraphics[width=0.4\textwidth]{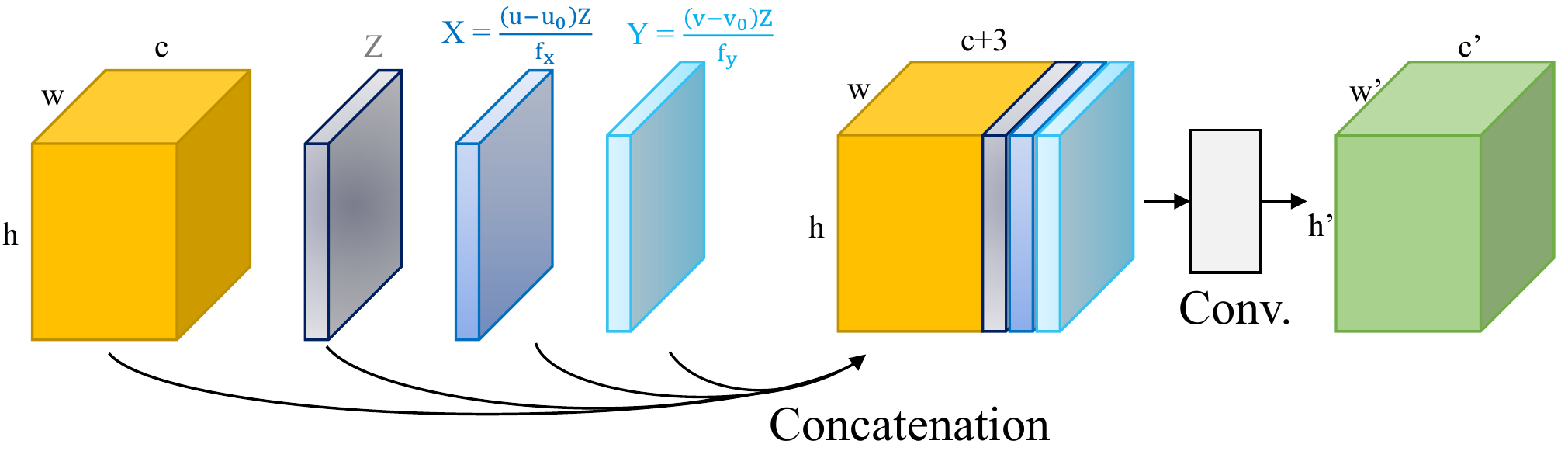}}
	\caption{Comparison of convolutional layers. A geometric convolutional layer augments a convolutional layer by concatenating three extra channels, including $X$, $Y$, and $Z$, to the input.	
	 }
	\label{fig_geoconv}
\end{figure}

\subsection{The Geometric Convolutional Layer} 
As pointed out by~\cite{Chen20192d3d}, 3D geometric clues are important for depth completion. In this work, we propose a geometric convolutional layer to encode the 3D geometric information. As shown in Figure~\ref{fig_geoconv}, it simply augments a conventional convolutional layer via concatenating a 3D position map to the layer's input. The position map $(X, Y, Z)$ is derived from an original sparse depth map by
\begin{equation}
Z = D, \ \ \ \ X = \frac{(u - u_0) Z}{f_x}, \ \ \ \ Y = \frac{(v - v_0) Z}{f_y},
\end{equation}
in which $(u,v)$ are the coordinates of a pixel and $u_0, v_0, f_x, f_y$ are intrinsic parameters of a camera.

In this work, we replace each convolutional layer within each of the ResBlocks by the proposed geometric convolutional layer. In addition, the sparse depth map is min-pooled to obtain $Z$ at smaller scales. By this means, 3D geometric information can be better encoded into features in both color- and depth-dominant branches.

%
%
%
%
%
%
%
%



\subsection{The Dilated and Accelerated CSPN++}
\label{sec:DA-CSPN++}

As shown in~\cite{Cheng2018CSPN}, depth maps produced by a deep neural network may not preserve the input depth values at valid pixels. To recover the depth values at valid pixels, we adopt CSPN++~\cite{Cheng2020CSPN++} to refine the depth map predicted by our backbone. Based on CSPN++, we design two modifications to make it more effective and efficient. First, we introduce a dilation strategy similar to the well known dilated convolutions~\cite{Yu2016} to enlarge the propagation neighborhoods. Second, we design an implementation that makes the propagation from each neighbor truly parallel, which greatly accelerates the propagation procedure.

%


Hereby, we briefly introduce our implementation for acceleration. Formally, we denote a coarse depth map by $D^0$. The spatial propagation network produces a refined depth map $D^t$ after $t$ iterations. For pixel $\mathbf{i}$, at each iteration it aggregates information propagated from pixels within its neighborhood $\mathcal{N}(\mathbf{i})$. That is,
\begin{equation}
		\label{eq:SPN}
	D_\mathbf{i}^{t+1}=W_{\mathbf{i}\mathbf{i}}D_\mathbf{i}^0+\sum_{\mathbf{j}\in \mathcal{N}(\mathbf{i})}W_{\mathbf{j}\mathbf{i}}D_\mathbf{j}^t,
\end{equation}
where $W_{\mathbf{j}\mathbf{i}}$ denotes the affinity between pixel $\mathbf{i}$ and pixel $\mathbf{j}$.

\begin{figure}[ht]
	\centering
	\includegraphics[width=0.495\textwidth]{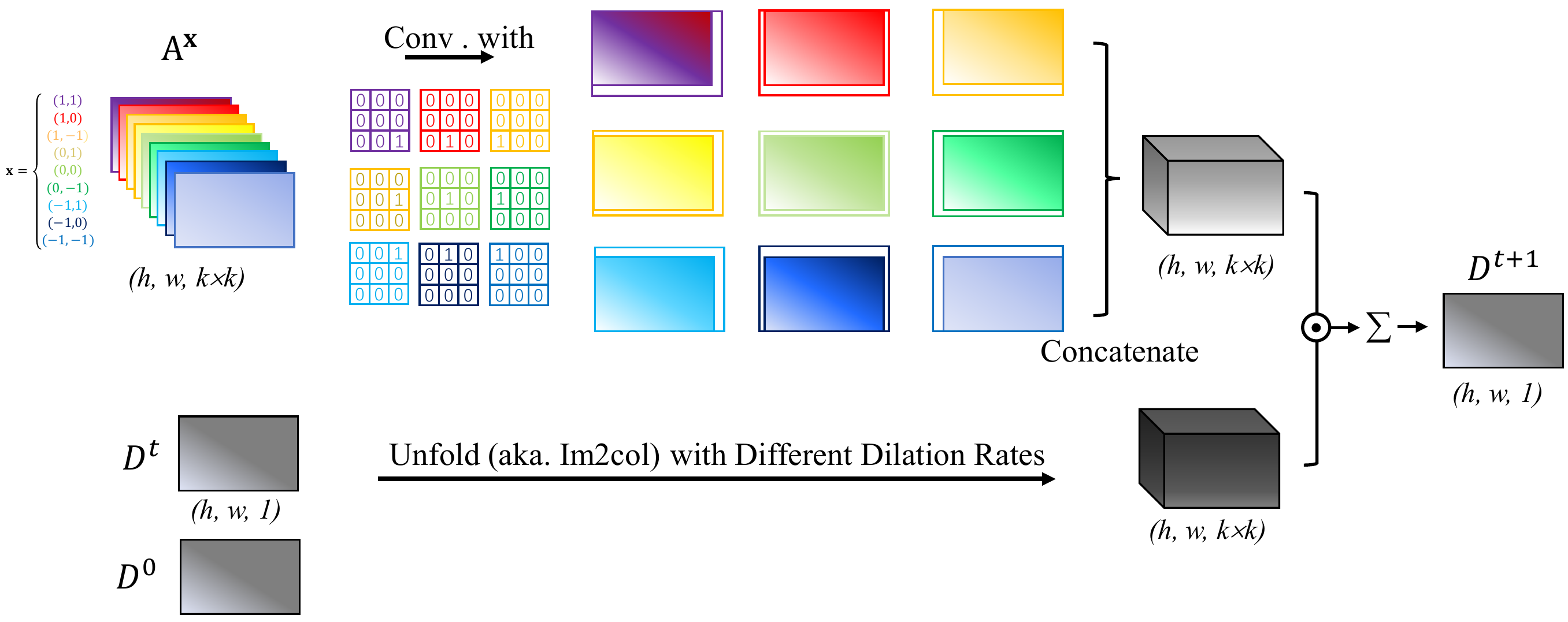}
	\caption{Illustration of our accelerated implementation. Each affinity map is translated by convolving with a corresponding one-hot convolutional kernel. The propagation is then conducted at tensor-level and fully parallel.}
	\label{fig_DACSPN}
\end{figure}

This equation is defined pixel-wise. For the purpose of efficiency, we convert it to a tensor-level operation. Considering a neighborhood of $k\times k$ size, from the network we learn $k\times k$ number of affinity maps, each of which represents the affinity of one certain neighbor to all pixels. 
Then, each affinity map needs to be translated along the opposite direction of the corresponding neighbor for alignment. Taking a $3\times 3$ neighborhood as an example, we use nine one-hot convolutional kernels to implement these translations, as shown in Figure~\ref{fig_DACSPN}. We denote a translation operator by $\mathcal{T}(A^\textbf{x}, \textbf{x})$, which moves an affinity map $A^\textbf{x}$ along $-\textbf{x}$ direction. Then, the spatial propagation defined in Equation (\ref{eq:SPN}) is equivalent to the following one:
\begin{equation}
\label{eq:spn2}
D^{t+1}=
\mathcal{T}(A^\mathbf{0}, \mathbf{0})
\mathcal{T}(D^0,\mathbf{0})+
\sum_{\mathbf{x}\in {\mathcal{N}}}
\mathcal{T}(A^{\mathbf{x}}, \mathbf{x})
\mathcal{T}(D^t,\mathbf{x})
\end{equation}

By using one-hot convolutional kernels, our implementation of the translations can be performed parallelly. Moreover, the transformed propagation defined in Equation (\ref{eq:spn2}) can be implemented much more efficient than the pixel-wise one.



\begin{figure*}[ht]
	\centering
	\vspace{0cm}
	\subfigtopskip=3pt
	\subfigbottomskip=-10pt
	\rotatebox{90}{\scriptsize \textcolor{white}{p} \  RGB}
	\subfigure[]{\includegraphics[width=0.3\linewidth]{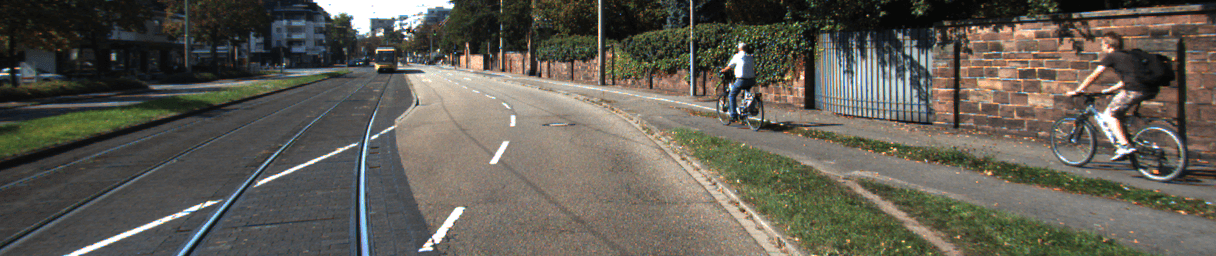}}
	\subfigure[]{\includegraphics[width=0.3\linewidth]{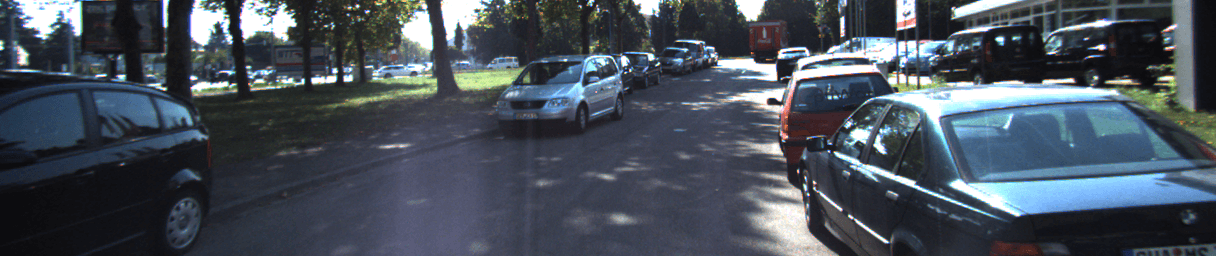}}
	\subfigure[]{\includegraphics[width=0.3\linewidth]{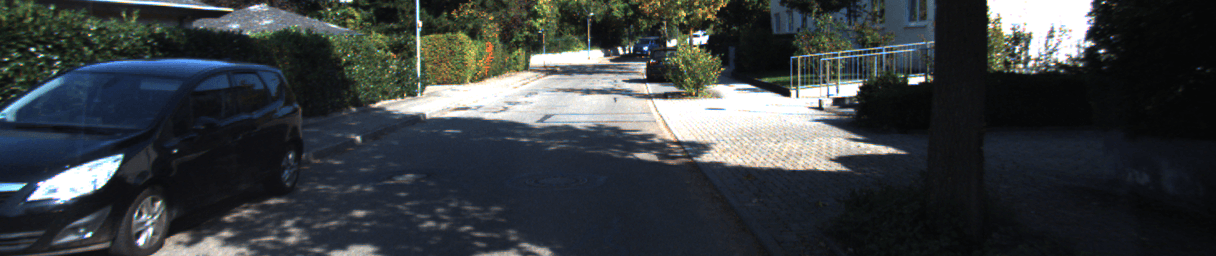}}

	\rotatebox{90}{\scriptsize \textcolor{white}{p}\quad SD}
	\subfigure[]{\includegraphics[width=0.3\linewidth]{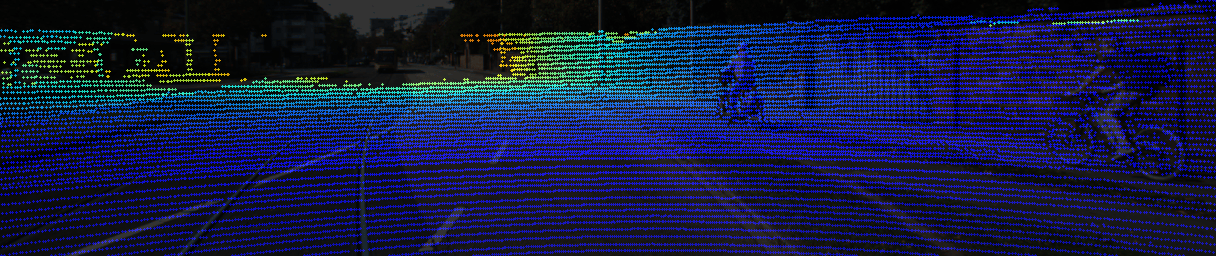}}
	\subfigure[]{\includegraphics[width=0.3\linewidth]{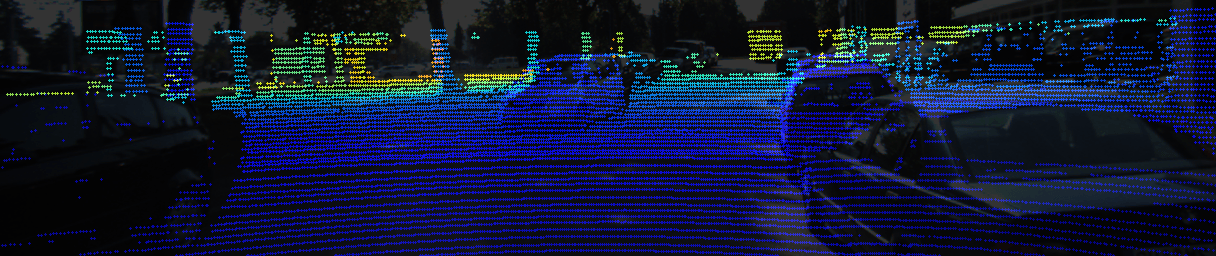}}
	\subfigure[]{\includegraphics[width=0.3\linewidth]{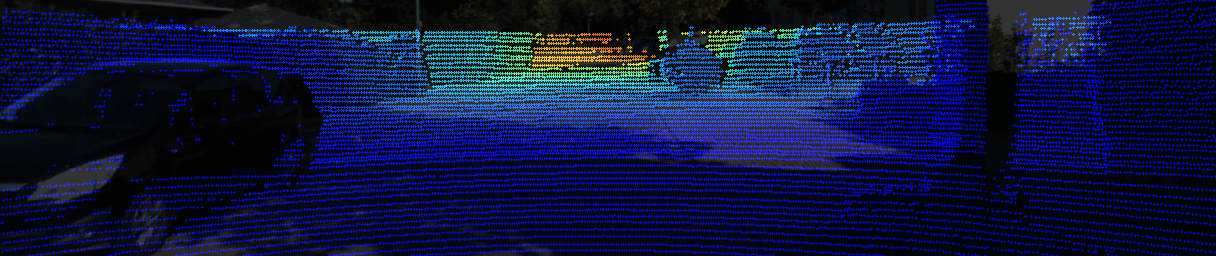}}
	
	\rotatebox{90}{\scriptsize \textcolor{white}{p}\quad GT}
	\subfigure[]{\includegraphics[width=0.3\linewidth]{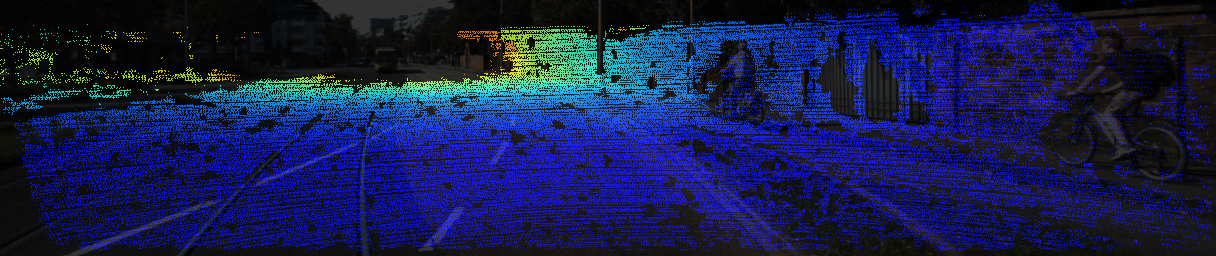}}
	\subfigure[]{\includegraphics[width=0.3\linewidth]{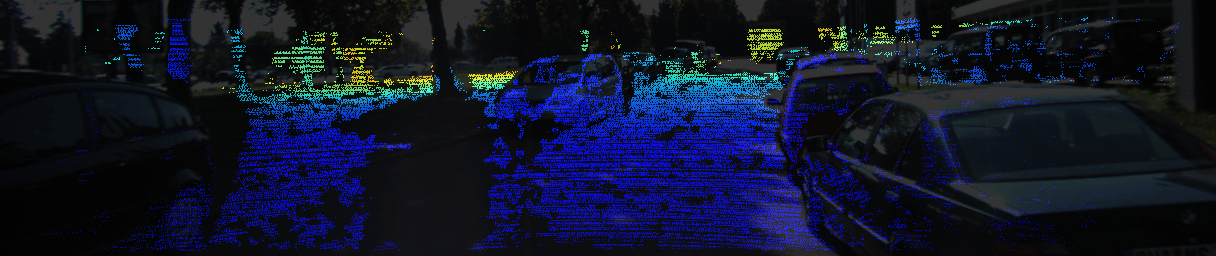}}
	\subfigure[]{\includegraphics[width=0.3\linewidth]{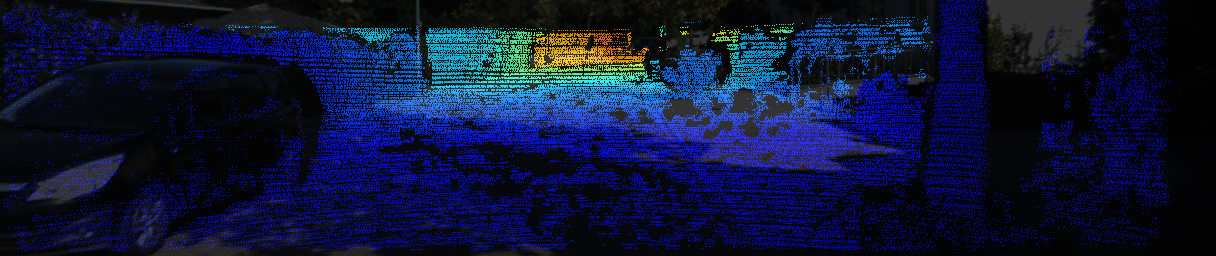}}
	
	\rotatebox{90}{\scriptsize \textcolor{white}{p}CD-Dep.}
	\subfigure[]{\includegraphics[width=0.3\linewidth]{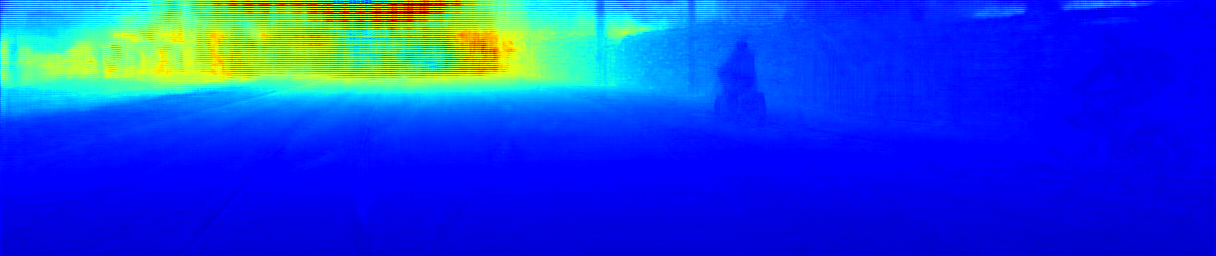}}
	\subfigure[]{\includegraphics[width=0.3\linewidth]{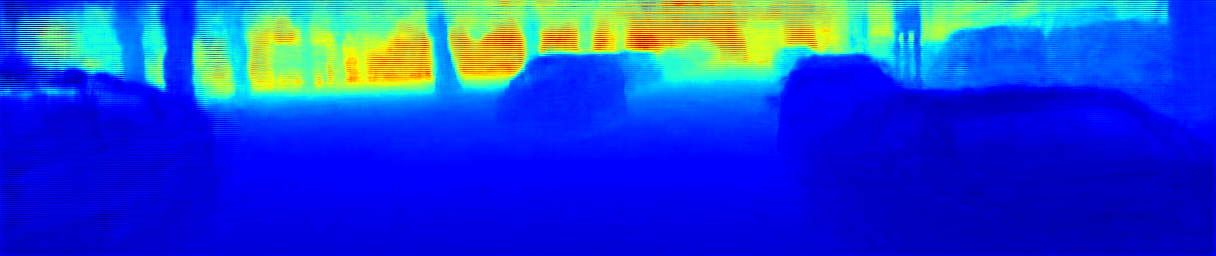}}
	\subfigure[]{\includegraphics[width=0.3\linewidth]{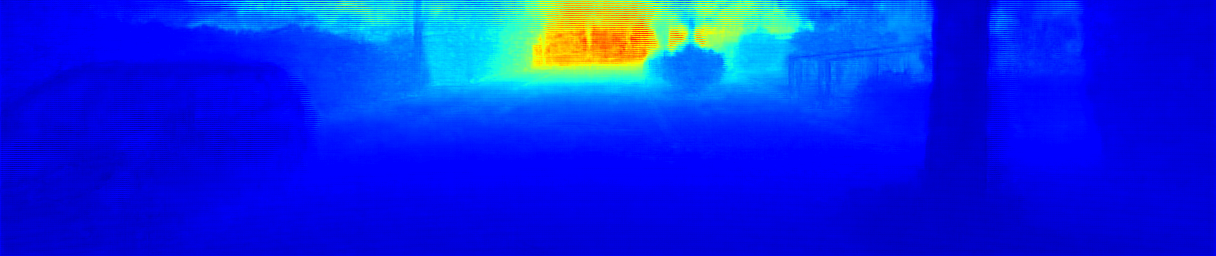}}
	
	\rotatebox{90}{\scriptsize \textcolor{white}{p}DD-Dep.}
	\subfigure[]{\includegraphics[width=0.3\linewidth]{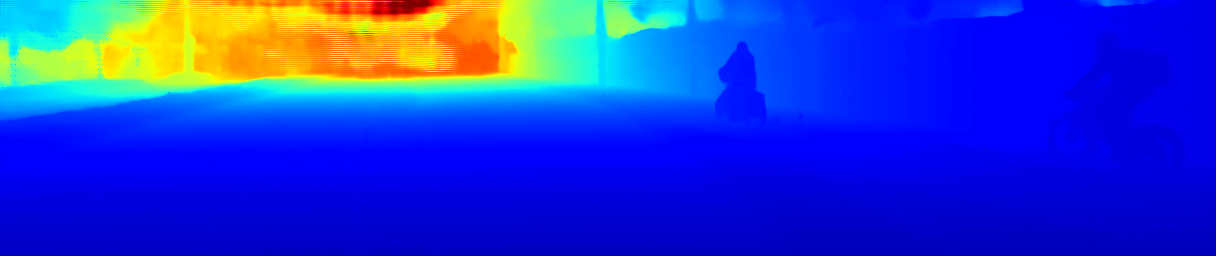}}
	\subfigure[]{\includegraphics[width=0.3\linewidth]{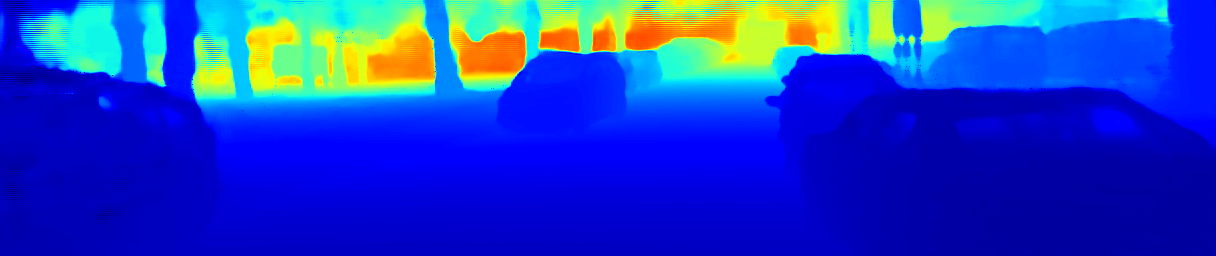}}
	\subfigure[]{\includegraphics[width=0.3\linewidth]{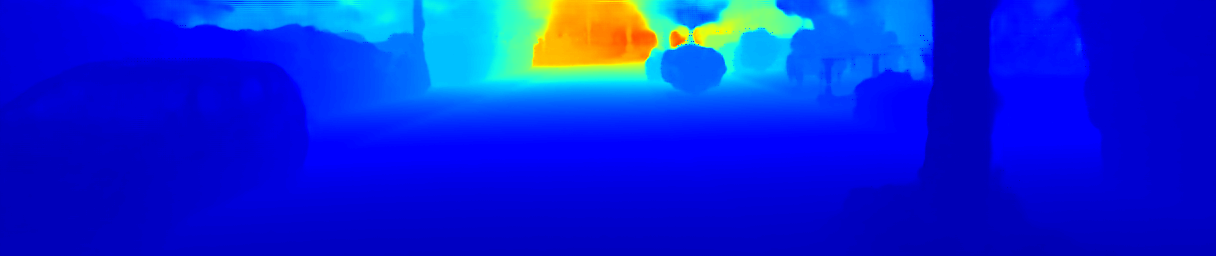}}
	
	\rotatebox{90}{\scriptsize \textcolor{white}{p} \ Fused}
	\subfigure[]{\includegraphics[width=0.3\linewidth]{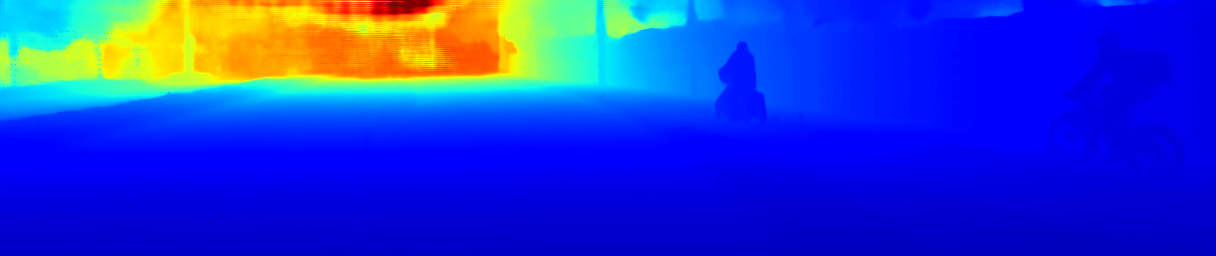}}
	\subfigure[]{\includegraphics[width=0.3\linewidth]{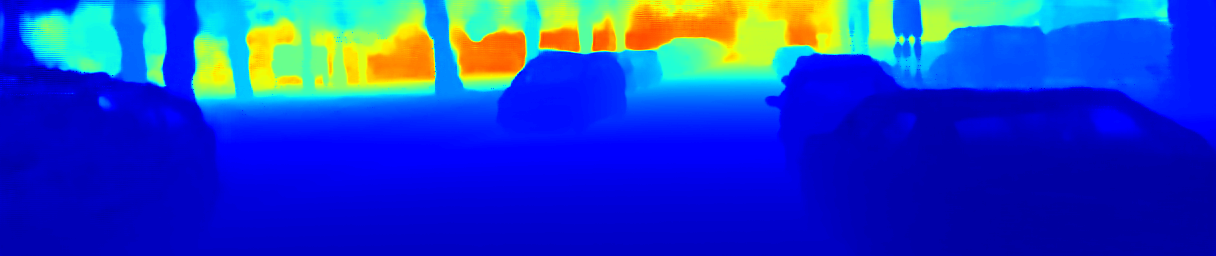}}
	\subfigure[]{\includegraphics[width=0.3\linewidth]{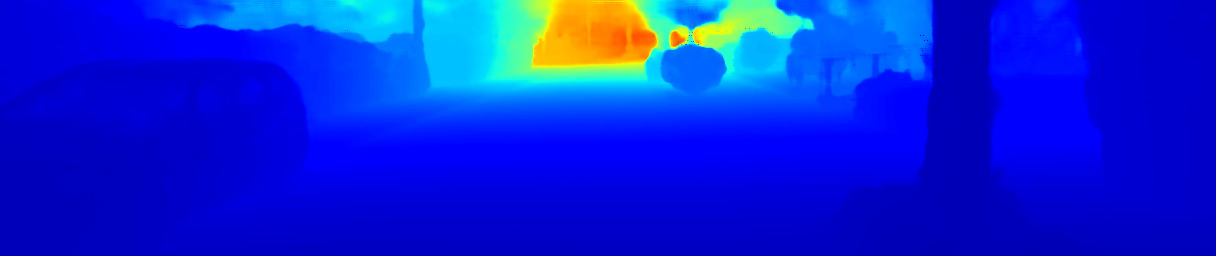}}
	
	\rotatebox{90}{\scriptsize \textcolor{white}{p}\ Refined}
	\subfigure[]{\includegraphics[width=0.3\linewidth]{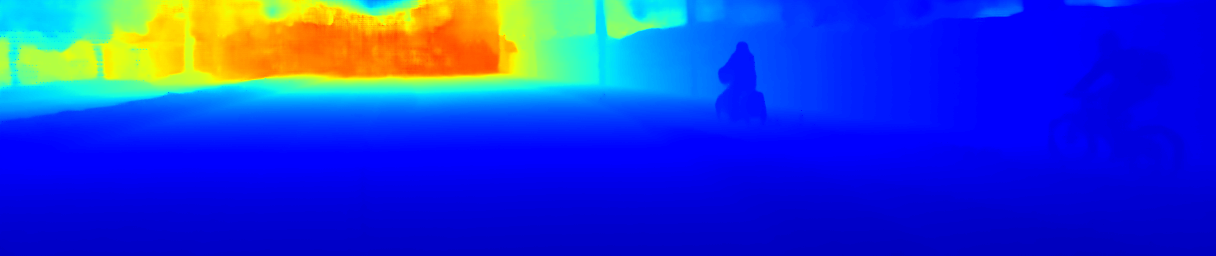}}
	\subfigure[]{\includegraphics[width=0.3\linewidth]{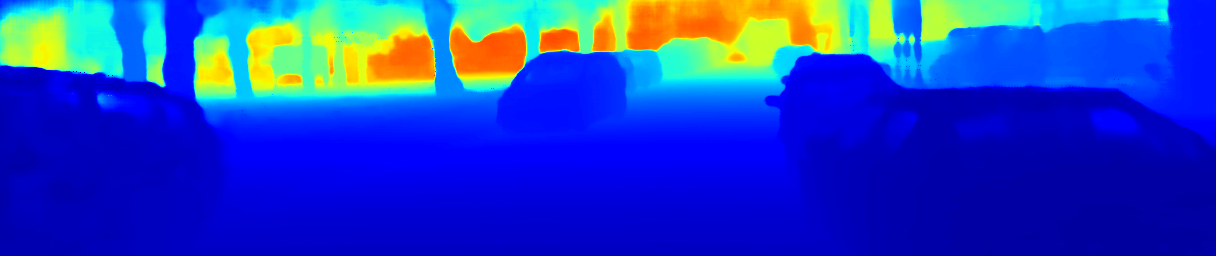}}
	\subfigure[]{\includegraphics[width=0.3\linewidth]{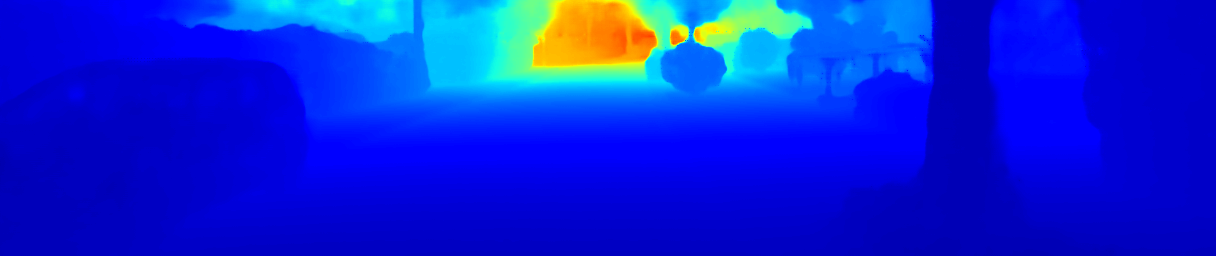}}
	
	\rotatebox{90}{\scriptsize \textcolor{white}{p}CD-Conf.}
	\subfigure[]{\includegraphics[width=0.3\linewidth]{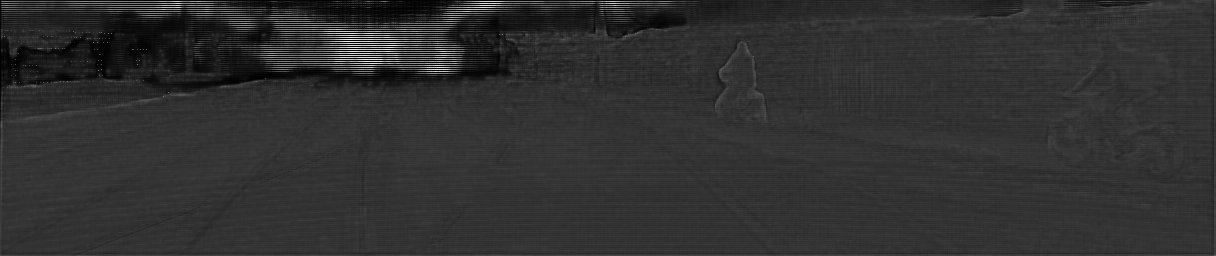}}
	\subfigure[]{\includegraphics[width=0.3\linewidth]{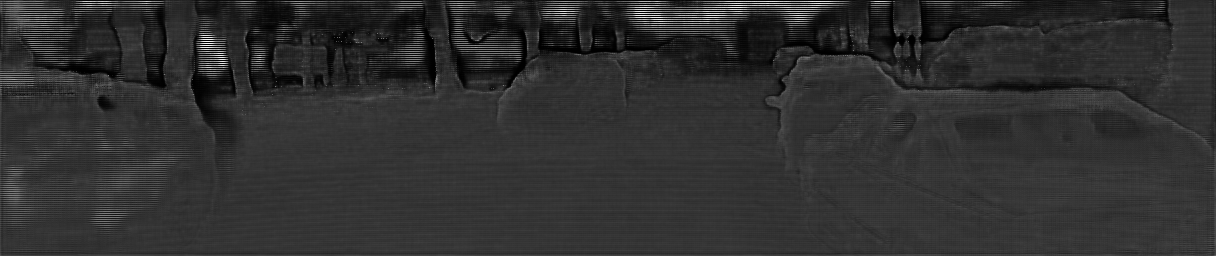}}
	\subfigure[]{\includegraphics[width=0.3\linewidth]{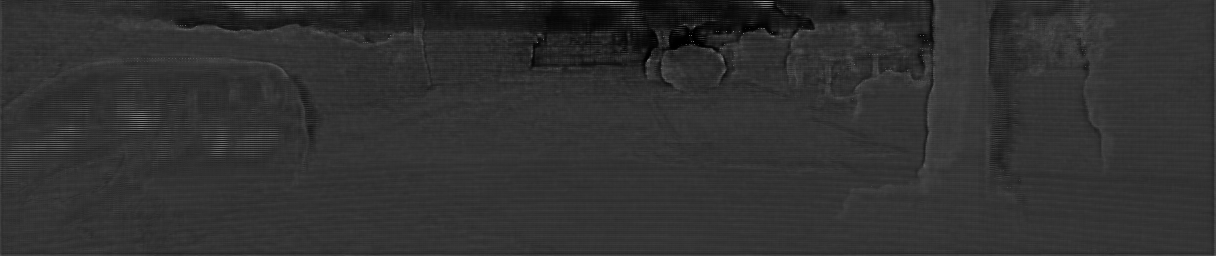}}
	
	\rotatebox{90}{\scriptsize \textcolor{white}{p}DD-Conf.}
	\subfigure[]{\includegraphics[width=0.3\linewidth]{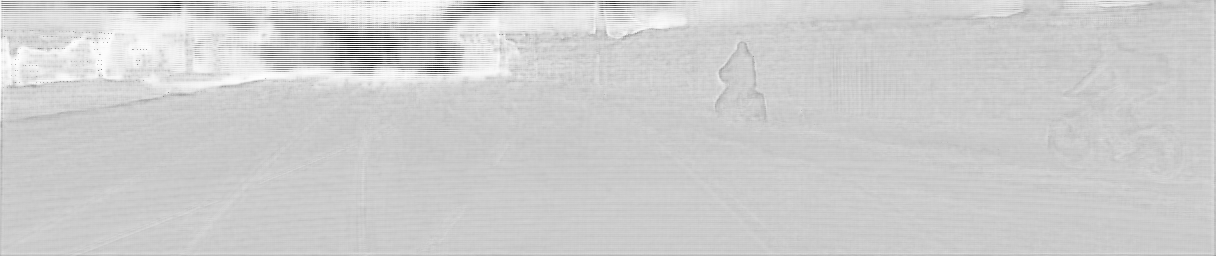}}
	\subfigure[]{\includegraphics[width=0.3\linewidth]{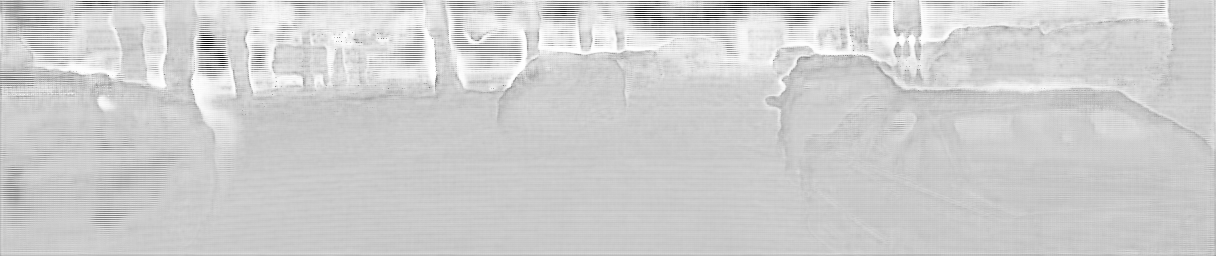}}
	\subfigure[]{\includegraphics[width=0.3\linewidth]{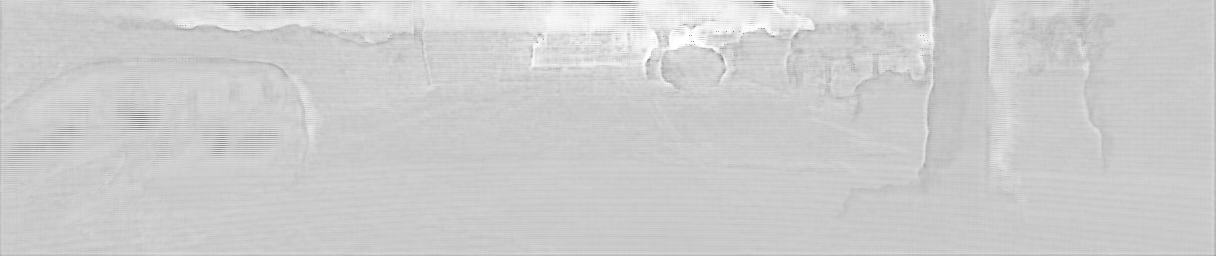}}
	
	\caption{Illustrations of typical examples. The outputs, including CD-Depth, DD-Depth, CD-Confidence, DD-Confidence, and fused depth maps, are obtained by the geometric encoded backbone (the model B$_4$+GCL). We also provide the refined depth maps obtained by our full model B$_4$+GCL+C2 for reference.}
	\label{fig_results}
\end{figure*}

\begin{table*}[htbp]
	\begin{center}
		\begin{tabular}{c|ccccc|cccc}
		\hline
			Models    & \begin{tabular}[c]{@{}c@{}}CD-Input\\ Sparse Depth\end{tabular} & \begin{tabular}[c]{@{}c@{}}DD-Input\\ CD-Depth \end{tabular} & \begin{tabular}[c]{@{}c@{}}CD-Output \\ Guidance Map \end{tabular} & \begin{tabular}[c]{@{}c@{}}Geometric \\ Encoding\end{tabular} & \begin{tabular}[c]{@{}c@{}}CSPN++ \\ Dilation\end{tabular} & RMSE & MAE & iRMSE & iMAE  \\ \hline
			B$_1$ & & & & & & 839.11 & 285.27 & 2.77 & 1.44 \\
			B$_2$ & $\surd$ & & & &  & 824.61 & 271.37 & 5.09 & 1.51 \\
			B$_3$ & & $\surd$ & & & & 802.29 & 227.18 & 2.26 & 0.99 \\
			B$_4$ & $\surd$ & $\surd$ & & & & 781.66 & 215.57 & 2.21 & 0.94 \\
			B$_5$ & $\surd$ & $\surd$ & $\surd$ & & & 783.91 & 220.01 & 2.24 & 0.97 \\ \hline
			B$_4$+CCL & $\surd$ & $\surd$ & & u,v & & 783.19 & 226.29 & 2.63 & 0.99 \\
			B$_4$+DCL & $\surd$ & $\surd$ & &  Z & & 777.89 & 216.04 & 2.21 & 0.94 \\
			B$_4$+GCL & $\surd$ & $\surd$ &  & X,Y,Z & &772.78 & 215.48 & 2.18 & 0.94 \\ \hline
			B$_4$+C1  & $\surd$ & $\surd$ & & &1 & 765.83 & 212.99 & 2.28 & 0.97 \\
			B$_4$+C2 & $\surd$ & $\surd$  & & &2,1 & 762.84 & 209.28 & 2.19 & 0.92 \\
			B$_4$+C4 & $\surd$ & $\surd$ & & &4,2,1 & 763.60 & 212.71 & 2.26 & 0.95 \\ \hline
			B$_4$+GCL+C2 & $\surd$ & $\surd$ & & X,Y,Z & 2,1 & 757.20 & 209.00 & 2.22 & 0.92 \\ \hline
		\end{tabular}
	\end{center}
	\caption{Performance on the KITTI depth completion validation set.}
	\label{table:val}
\end{table*}


\subsection{The Training Loss}
We use a $\ell_2$ loss for training, which is defined by
\begin{equation}
	L(\hat{D}) = \left \| (\hat{D}-D_{gt})\odot \mathbbm{1}(D_{gt}>0) \right \|^2.
\end{equation}
Here, $\hat{D}$ is the predicted depth map, $D_{gt}$ is a ground truth for supervision, $\mathbbm{1}()$ is an indicator, and $\odot$ is an element-wise multiplication. Since the ground truth contains invalid pixels, we only consider those having valid depth values. 

In the early epochs of training, supervision is also placed to the intermediate depth prediction results. That is,
\begin{equation}
	\label{eq:intsup}
	L =  L(\hat{D}) + \lambda_{cd} L(\hat{D}_{cd}) + \lambda_{dd} L(\hat{D}_{dd}),
\end{equation}
where $\lambda_{cd}$ and $\lambda_{dd}$ are two hyper-parameters empirically set.

\section{Experiments}
\subsection{Experimental Setup}
\textit{Dataset:} We evaluate the proposed model and its variants on the KITTI depth completion dataset~\cite{KITTI,Uhrig2017sparsity}. It provides both color images and aligned sparse depth maps that are obtained by projecting 3D LiDAR points to corresponding image frames. The images are in the resolution of $1216 \times 352$. A sparse depth map has about $5\%$ valid pixels and a ground truth dense depth map have around $16\%$ valid pixels~\cite{Uhrig2017sparsity}. The dataset contains 86K frames for training, together with 7K validation frames and 1K test frames. In the validation set, 1K frames are officially selected~\cite{KITTI,Uhrig2017sparsity}. We use the 1K validation set for ablation studies. 
%

%
%

\textit{Evaluation metrics:} As the common practice, we adopt four metrics for performance evaluation, which are root mean squared error (RMSE [mm]), mean absolute error (MAE [mm]), root mean squared error of the inverse depth (iRMSE [1/km]), and mean absolute error of the inverse depth (iMAE [1/km]). Besides, the runtime of inference is also reported. 



\textit{Implementation details:} The proposed model is implemented with the PyTorch~\cite{PyTorch} framework and trained on two NVIDIA GTX 2080Ti GPUs. During training, we use the ADAM optimizer~\cite{adam} with $\beta_1=0.9$, $\beta_2=0.99$, and the weight decay is $10^{-6}$. We adopt a multi-stage training strategy to train the backbone, DA-CSPN++, and the full model progressively. First, the backbone is trained with a batch size of 6 and an initial learning rate of 0.001, decayed by \{$\frac{1}{2}$, $\frac{1}{5}$, $\frac{1}{10}$\} at epoch \{10, 15, 25\}, for 30 epochs. The loss defined in Equation (\ref{eq:intsup}) is used, with $\lambda_{cd} = \lambda_{dd} = 0.2$ at initial epochs and reduced to 0 later. Then, we freeze the weights in the backbone and train DA-CSPN++ with a batch size of 6 and a learning rate of of 0.001 for 2 epochs. Finally, we train the full model with an initial learning rate of 0.02 and 0.002, respectively, for the weights in the backbone and DA-CSPN++. In this stage, the training procedure lasts for 75 epochs, and the learning rate is decayed by \{$\frac{1}{2}$, $\frac{1}{5}$, $\frac{1}{10}$, $\frac{1}{20}$, $\frac{1}{50}$\} at epoch \{10, 20, 30, 40, 50\}. We randomly crop images to $576\times 160$ and set the batch size to 10, making it feasible to train effectively with limited computational resources. In addition, data augmentation techniques including horizontal random flip and color jitter~\cite{s2d} are adopted.

\subsection{Ablation Studies}
We first conduct a series of experiments to validate the effectiveness of each component proposed in our method, including the two-branch backbone, the geometric convolutional layer, and the DA-CPSN++ module.
%
%

\textit{The effectiveness of the two-branch backbone.} According to whether input a sparse depth map into the color-dominant branch and input the CD-depth prediction into the depth-dominant branch or not, we obtain four variants of the backbone. The performance of these variants, denoted by B$_1$ to B$_4$, are presented in Table~\ref{table:val}. The results show that the performance is greatly improved when the CD branch is assisted with the sparse depth input and the DD branch is guided with the CD-depth prediction. In addition, we investigate one more backbone variant (B$_5$), which additionally produces a guidance map from the first branch to guide the second one, as done in FusionNet~\cite{Gansbeke2019} and DeepLiDAR~\cite{Qiu2019deeplidar}. The results show that this additional guidance map is not necessary and it even slightly hurts the performance. 


Figure~\ref{fig_results} illustrates some typical examples. From the CD- and DD-depth predictions and their confidence maps, we make the following observations. 1) Overall, DD-depth maps contribute more to fused depth maps in most regions. 2) The CD-depth predictions rely heavily on color information so that they are sensitive to the change of color or texture, as shown in road markings, grass and tree leaves. In these regions, CD-depth predictions have even lower confidences. 3) Color images have sharp object boundaries while the input sparse depth maps are noisy around object boundaries. These result in CD-depth predictions having higher confidence along object boundaries than other regions.



\begin{figure}[th]
	\centering
	\vspace{0cm}
	\subfigtopskip=3pt
	\subfigbottomskip=-10pt
	\rotatebox{90}{\quad  \ RGB}
	\subfigure[]{\includegraphics[width=0.8\linewidth]{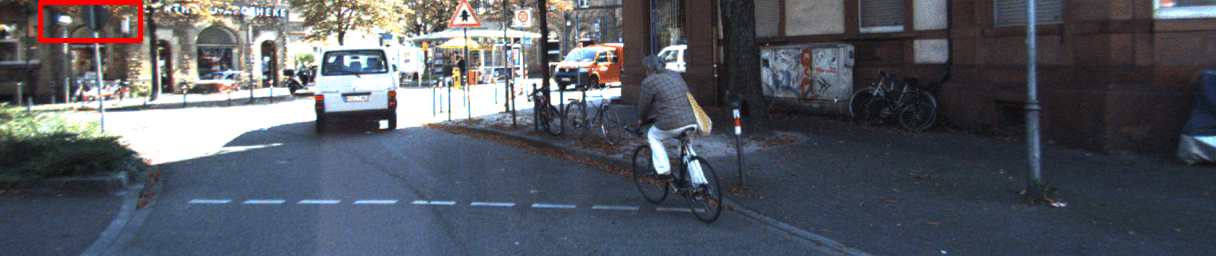}}
	
	\rotatebox{90}{\quad  \ \, SD}
	\subfigure[]{\includegraphics[width=0.8\linewidth]{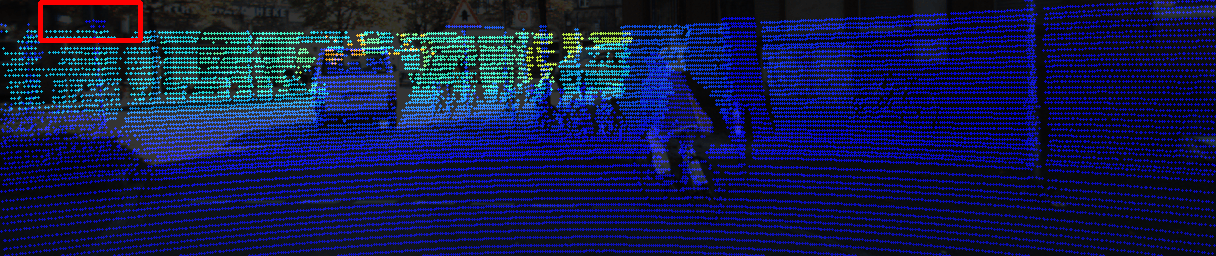}}
	
	\rotatebox{90}{\quad  \ \, GT}
	\subfigure[]{\includegraphics[width=0.8\linewidth]{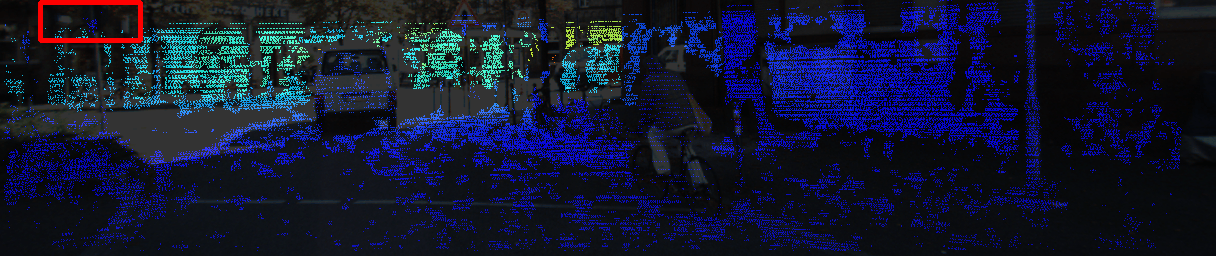}}
	
	\rotatebox{90}{\quad  \ \, CL}
	\subfigure[]{\includegraphics[width=0.8\linewidth]{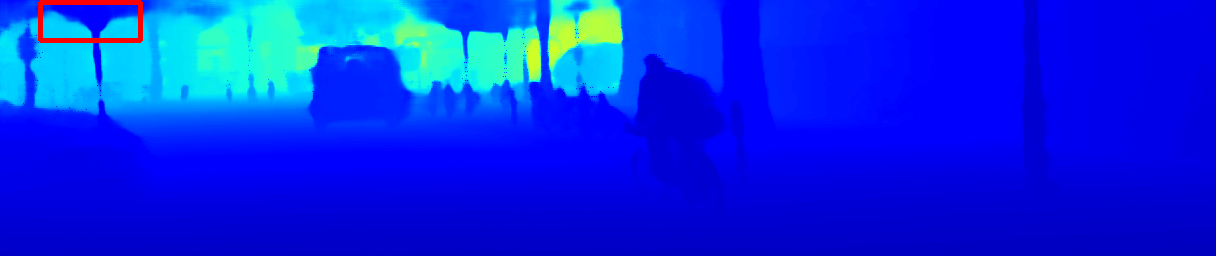}}
	
	\rotatebox{90}{\quad  \ CCL}
	\subfigure[]{\includegraphics[width=0.8\linewidth]{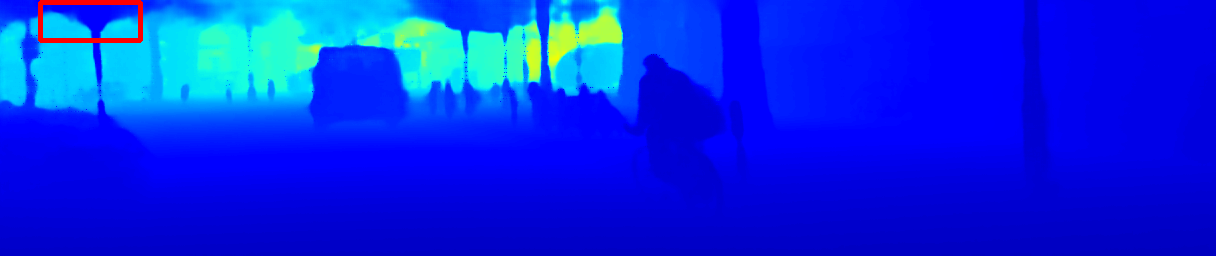}}
	
	\rotatebox{90}{\quad  \ DCL}
	\subfigure[]{\includegraphics[width=0.8\linewidth]{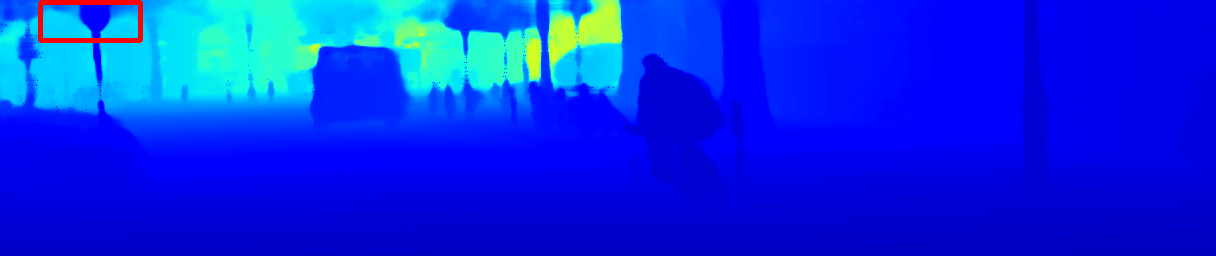}}
	
	\rotatebox{90}{\quad \ GCL}
	\subfigure[]{\includegraphics[width=0.8\linewidth]{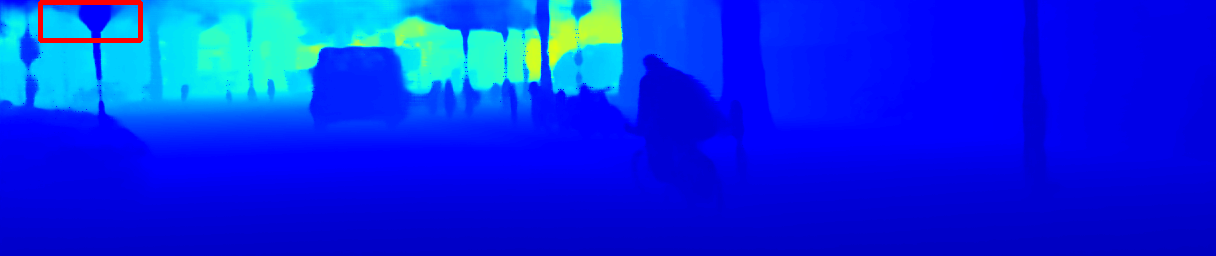}}
	
	\caption{A typical example to illustrate the difference of various encoding strategies. Compared to standard convolution(CL) and CoordConv(CCL), our geometric convolutional layer (GCL) infers better depth information.}
	\label{fig_gcl}
\end{figure}

\textit{The effectiveness of the geometric convolutional layer.} Based on the backbone model B$_4$, we further replace each convolutional layer in the ResBlocks by our proposed geometric convolutional layer and get the model B$_4$+GCL. Besides, we compare this component to the CoordConv layer~\cite{Liu2018Coordconv} that encodes pixel coordinates, together with another variant that encodes the depth only. These two variants are denoted as B$_4$+CCL and B$_4$+DCL respectively. As shown in Table~\ref{table:val}, our geometric convolutional layer improves the backbone's performance on RMSE by a great margin. B$_4$+DCL also helps for performance. However, encoding pixel coordinates (B$_4$+CCL) may slightly hurt the performance, contradicting to its performance in other position sensitive vision tasks~\cite{Wang2020SOLO,Choi2020fly}.

Figure~\ref{fig_gcl} demonstrates a typical example to show the difference of model B$_4$+GCL and its counterparts. The model with geometric convolutional layers can infer better depth information than the other models especially when a foreground object looks similar to background in color, as the road sign shown in marked red boxes.

\textit{The effectiveness of DA-CSPN++.} Based on the backbone model B$_4$, we integrate variants of CSPN++ to compare their performance. The total number of iterations for propagation is 12. C1 stands for original CSPN++, with a dilation rate ($dr$) of 1 for all iterations. C2 stands for the model that takes $dr=2$ for first six iterations and $dr=1$ for the remaining iterations. C4 is the model taking $dr=\{4, 2, 1\}$ for every four iterations, respectively. As shown in Table~\ref{table:val}, all variant models can greatly improve the backbone's performance and the model B$_4$+C2 slightly outperforms the other two counterparts. Table~\ref{table:runtime} lists the time taken by the spatial propagation in the depth refinement module, tested on our single 2080Ti GPU. The results show that our accelerated implementation greatly reduces the running time. 


%
%

\begin{table}[htbp]
	\begin{center}
		\begin{tabular}{c|c|c}
		\hline 
			SPN Models & Acceleration & Propagation Time \\ \hline 
			NLSPN\cite{Park2020NLSPN} & \textemdash &0.055s \\ 
			C1\cite{Cheng2020CSPN++} & & 0.091s\\ 
			C2 & & 0.186s \\
			C1 & $\surd$ & 0.014s\\ 
			C2 & $\surd$ & 0.015s \\ \hline
		\end{tabular}
	\end{center}
	\caption{Runtime of the depth refinement modules.}
	\label{table:runtime}
\end{table}

\subsection{Comparison with State-of-the-arts}
The proposed method ranks 1st in the KITTI online leaderboard at the time of submission. In table~\ref{tab:sota}, we present the quantitative performance of our full method (referred to as PENet), together with the other top 10 methods that have published or archived papers. The results show that our method has a significant improvement on RMSE, which is the most important metric for evaluation. We also test our geometric encoded backbone without depth refinement (referred to as ENet). The results show that this model outperforms 9 top ranked methods, including all of these~\cite{Park2020NLSPN,Cheng2020CSPN++,Xu2020dspn} using spatial propagation techniques.

Table~\ref{tab:sota} also provides two runtimes. Runtime1 is quoted from the leaderboard. To make a fair comparison, Runtime2 is tested on our single 2080Ti GPU with source codes released by the authors. The results indicate that our full model infers faster than 8 methods. Especially, it runs much faster than those~\cite{Park2020NLSPN,Cheng2020CSPN++,Xu2020dspn} that utilize spatial propagation techniques as well.

\begin{table}[htbp]
	\begin{center}
		\resizebox{.49\textwidth}{!}{
			\begin{tabular}{l|c c c c|c|c}
				\hline
				Method  & RMSE & MAE & iRMSE & iMAE & Runtime1 & Runtime2\\ \hline
				PENet(Ours) & 730.08 & 210.55 & 2.17 & 0.94 & 0.04s & 0.032s \\
				GuideNet~\cite{Tang2019guided}  & 736.24 & 218.83 & 2.25 & 0.99 & 0.14s & - \\
				ENet(Ours) & 741.30 & 216.26 & 2.14 & 0.95 & 0.02s & 0.019s \\
				NLSPN~\cite{Park2020NLSPN}  & 741.68 & 199.59 & 1.99 & 0.84 & 0.22s & 0.127s \\
				CSPN++~\cite{Cheng2020CSPN++}  & 743.69 & 209.28 & 2.07 & 0.90 & 0.2s & - \\
				ACMNet~\cite{Zhao2020context} & 744.91 & 206.09 & 2.08 & 0.90 & 0.2s & 0.330s \\
				UberATG~\cite{Chen20192d3d} & 752.88 & 221.19 & 2.34 & 1.14 & 0.09s & - \\
				DeepLiDAR~\cite{Qiu2019deeplidar}  & 758.38 & 226.50 & 2.56 & 1.15 & 0.07s & 0.051s\\
				MSG-CHN~\cite{Li2020MSG} & 762.19 & 220.41 & 2.30 & 0.98 & 0.01s & 0.011s \\
				DSPN~\cite{Xu2020dspn} & 766.74 & 220.36 & 2.47 & 1.03 & 0.34s & -\\
				FusionNet~\cite{Gansbeke2019} & 772.87 & 215.02 & 2.19 & 0.93 & 0.02s & 0.022s  \\
				PwP~\cite{Xu2019} & 777.05 & 235.17 & 2.42 & 1.13 & 0.1s &- \\ 
				\hline
		\end{tabular}}	
	\end{center}
	\caption{Comparisons to state-of-the-art methods on the KITTI test set, ranked by RMSE.}
	\label{tab:sota}
\end{table}

\section{Conclusion}
In this paper, we have presented a method for image guided depth completion. By revisiting two-branch architectures developed in previous works, we propose a new two-branch architecture that exploit color- and depth-dominant information, respectively, from two branches. The designed backbone, together with the proposed geometric convolutional layer, can exploit and fuse multi-modality information thoroughly. In addition, we integrate a speedup DA-CSPN++ module for further depth refinement. The entire model is precise and efficient, as tested in the KITTI online leaderboard.

\addtolength{\textheight}{-8cm}   

\bibliographystyle{IEEEtran}
\bibliography{PEDC}

\begin{thebibliography}{10}
\providecommand{\url}[1]{#1}
\csname url@rmstyle\endcsname
\providecommand{\newblock}{\relax}
\providecommand{\bibinfo}[2]{#2}
\providecommand\BIBentrySTDinterwordspacing{\spaceskip=0pt\relax}
\providecommand\BIBentryALTinterwordstretchfactor{4}
\providecommand\BIBentryALTinterwordspacing{\spaceskip=\fontdimen2\font plus
\BIBentryALTinterwordstretchfactor\fontdimen3\font minus
  \fontdimen4\font\relax}
\providecommand\BIBforeignlanguage[2]{{%
\expandafter\ifx\csname l@#1\endcsname\relax
\typeout{** WARNING: IEEEtran.bst: No hyphenation pattern has been}%
\typeout{** loaded for the language `#1'. Using the pattern for}%
\typeout{** the default language instead.}%
\else
\language=\csname l@#1\endcsname
\fi
#2}}

\bibitem{Jaritz2018semantic}
M.~Jaritz, R.~de~Charette, E.~Wirbel, X.~Perrotton, and F.~Nashashibi, ``Sparse
  and dense data with cnns: Depth completion and semantic segmentation,'' in
  \emph{3DV}, 2018.

\bibitem{Hua2018normalized}
J.~Hua and X.~Gong, ``A normalized convolutional neural network for guided
  sparse depth upsampling,'' in \emph{IJCAI}, 2018.

\bibitem{Tang2019guided}
J.~Tang, F.-P. Tian, W.~Feng, J.~Li, and P.~Tan, ``Learning guided
  convolutional network for depth completion,'' \emph{arXiv preprint
  arXiv:1908.01238}, 2019.

\bibitem{Gansbeke2019}
W.~V. Gansbeke, D.~Neven, B.~D. Brabandere, and L.~V. Gool, ``Sparse and noisy
  lidar completion with rgb guidance and uncertainty,'' in \emph{MAV}, 2019.

\bibitem{Qiu2019deeplidar}
J.~Qiu, Z.~Cui, Y.~Zhang, X.~Zhang, S.~Liu, B.~Zeng, and M.~Pollefeys,
  ``Deeplidar: Deep surface normal guided depth prediction for outdoor scene
  from sparse lidar data and single color image,'' in \emph{CVPR}, 2019.

\bibitem{Cordts2016}
M.~Cordts, M.~Omran, S.~Ramos, T.~Rehfeld, M.~Enzweiler, R.~Benenson,
  U.~Franke, S.~Roth, and B.~Schiele, ``The cityscapes dataset for semantic
  urban scene understanding,'' in \emph{CVPR}, 2016.

\bibitem{Cheng2020CSPN++}
X.~Cheng, P.~Wang, C.~Guan, and R.~Yang, ``Cspn++: Learning context and
  resource aware convolutional spatial propagation networks for depth
  completion,'' in \emph{AAAI}, 2020.

\bibitem{Uhrig2017sparsity}
J.~Uhrig, N.~Schneider, L.~Schneider, U.~Franke, T.~Brox, and A.~Geiger,
  ``Sparsity invariant cnns,'' in \emph{3DV}, 2017.

\bibitem{Eldesokey2020uncertainty}
A.~Eldesokey, M.~Felsberg, K.~Holmquist, and M.~Persson, ``Uncertainty-aware
  cnns for depth completion: Uncertainty from beginning to end,'' in
  \emph{CVPR}, 2020.

\bibitem{Liu2013}
J.~Liu and X.~Gong, ``Guided depth enhancement via anisotropic diffusion,'' in
  \emph{PCM}, 2013.

\bibitem{Cheng2018CSPN}
X.~Cheng, P.~Wang, and R.~Yang, ``Depth estimation via affinity learned with
  convolutional spatial propagation network,'' in \emph{ECCV}, 2018.

\bibitem{Eldesokey2018}
A.~Eldesokey, M.~Felsberg, and F.~S. Khan, ``Propagating confidences through
  cnns for sparse data regression,'' in \emph{BMVC}, 2018.

\bibitem{Huang2019hms-net}
Z.~Huang, J.~Fan, S.~Cheng, S.~Yi, X.~Wang, and H.~Li, ``Hms-net: Hierarchical
  multi-scale sparsity-invariant network for sparse depth completion,''
  \emph{IEEE Transactions on Image Processing}, pp. 3429--3441, 2019.

\bibitem{Eigen2014}
D.~Eigen, C.~Puhrsch, and R.~Fergus, ``Depth map prediction from a single image
  using a multi-scale deep network,'' in \emph{NIPS}, 2014.

\bibitem{Wang2018multiscale}
B.~Wang, Y.~Feng, and H.~Liu, ``Multi-scale features fusion from sparse lidar
  data and single image for depth completion,'' \emph{Electronics Letters},
  vol.~54, no.~24, pp. 1375--1377, 2018.

\bibitem{Li2020MSG}
A.~Li, Z.~Yuan, Y.~Ling, W.~Chi, S.~Zhang, and C.~Zhang, ``A multi-scale guided
  cascade hourglass network for depth completion,'' in \emph{WACV}, 2020.

\bibitem{Xu2019}
Y.~Xu, X.~Zhu, J.~Shi, G.~Zhang, H.~Bao, and H.~Li, ``Depth completion from
  sparse lidar data with depth-normal constraints,'' in \emph{ICCV}, 2019.

\bibitem{Schneider2016semantic}
N.~Schneider, L.~Schneider, P.~Pinggera, U.~Franke, M.~Pollefeys, and
  C.~Stiller, ``Semantically guided depth upsampling,'' in \emph{GCPR}, 2016.

\bibitem{Park2020NLSPN}
J.~Park, K.~Joo, Z.~Hu, C.-K. Liu, and I.~S. Kweon, ``Non-local spatial
  propagation network for depth completion,'' in \emph{ECCV}, 2020.

\bibitem{Chen20192d3d}
Y.~Chen, B.~Yang, M.~Liang, and R.~Urtasun, ``Learning joint 2d-3d
  representations for depth completion,'' in \emph{ICCV}, 2019.

\bibitem{Zhao2020context}
S.~Zhao, M.~Gong, H.~Fu, and D.~Tao, ``Adaptive context-aware multi-modal
  network for depth completion,'' \emph{arXiv preprint arXiv:2008.10833}, 2020.

\bibitem{Liu2018Coordconv}
R.~Liu, J.~Lehman, P.~Molino, F.~P. Such, E.~Frank, A.~Sergeev, and
  J.~Yosinski, ``An intriguing failing of convolutional neural networks and the
  coordconv solution,'' in \emph{NIPS}, 2018.

\bibitem{Wang2020SOLO}
X.~Wang, T.~Kong, C.~Shen, Y.~Jiang, and L.~Li, ``Solo: Segmenting objects by
  locations,'' in \emph{ECCV}, 2020.

\bibitem{Choi2020fly}
S.~Choi, J.~T. Kim, and J.~Choo, ``Cars can’t fly up in the sky: Improving
  urban-scene segmentation via height-driven attention networks,'' in
  \emph{CVPR}, 2020.

\bibitem{Liu2014SPN}
S.~Liu, S.~D. Mello, J.~Gu, G.~Zhong, M.-H. Yang, and J.~Kautz, ``Learning
  affinity via spatial propagation networks,'' in \emph{NIPS}, 2014.

\bibitem{he2016deep}
K.~He, X.~Zhang, S.~Ren, and J.~Sun, ``Deep residual learning for image
  recognition,'' in \emph{CVPR}, 2016.

\bibitem{Yu2016}
F.~Yu and V.~Koltun, ``Multi-scale context aggregation by dilated
  convolutions,'' in \emph{ICLR}, 2016.

\bibitem{KITTI}
A.~Geiger, P.~Lenz, and R.~Urtasun, ``Are we ready for autonomous driving? the
  kitti vision benchmark suite,'' in \emph{CVPR}, 2012.

\bibitem{PyTorch}
A.~Paszke, S.~Gross, F.~Massa, A.~Lerer, J.~Bradbury, G.~Chanan, Z.~Killeen,
  T.and~Lin, N.~Gimelshein, L.~Antiga, A.~Desmaison, A.~K{\"{o}}pf, E.~Yang,
  Z.~DeVito, M.~Raison, A.~Tejani, S.~Chilamkurthy, B.~Steiner, L.~Fang,
  J.~Bai, and S.~Chintala, ``Pytorch: An imperative style, high-performance
  deep learning library,'' in \emph{NeuroIPS}, 2019.

\bibitem{adam}
D.~Kingma and J.~Ba, ``Adam: A method for stochastic optimization.''
  \emph{arXiv preprint arXiv:1412.6980}, 2014.

\bibitem{s2d}
F.~Mal and S.~Karaman, ``Sparse-to-dense: Depth prediction from sparse depth
  samples and a single image,'' in \emph{ICRA}, 2018.

\bibitem{Xu2020dspn}
Z.~Xu, H.~Yin, and J.~Yao, ``Deformable spatial propagation network for depth
  completion,'' \emph{arXiv preprint arXiv:2007.04251}, 2020.

\end{thebibliography}

\end{document}